\documentclass{article}

\usepackage{PRIMEarxiv}

\usepackage[utf8]{inputenc} 
\usepackage[T1]{fontenc}    
\usepackage{hyperref}       
\usepackage{url}            
\usepackage{booktabs}       
\usepackage{amsfonts, amsmath}       
\usepackage{nicefrac}       
\usepackage{microtype}      
\usepackage{lipsum}
\usepackage{fancyhdr}       
\usepackage{graphicx}       
\usepackage{multirow}
\usepackage{tabularx}
\newcolumntype{C}{>{\centering\arraybackslash}X}

\graphicspath{{media/}}     

\title{Monte Carlo Tree Search Satellite Scheduling Under Cloud Cover Uncertainty
}

\author{
  Justin Norman, François Rivest \\
  Department of Mathematics and Computer Science \\
  Royal Military College of Canada \\
  Kingston, Canada\\
  \texttt{francois.rivest@\{rmc-cmr.ca, mail.mcgill.ca\}} \\
}

\begin{document}
\maketitle

\begin{abstract}
Efficient utilization of satellite resources in dynamic environments remains a challenging problem in satellite scheduling. This paper addresses the multi-satellite collection scheduling problem (m-SatCSP), aiming to optimize task scheduling over a constellation of satellites under uncertain conditions such as cloud cover. Leveraging Monte Carlo Tree Search (MCTS), a stochastic search algorithm, two versions of MCTS are explored to schedule satellites effectively. Hyperparameter tuning is conducted to optimize the algorithm's performance. Experimental results demonstrate the effectiveness of the MCTS approach, outperforming existing methods in both solution quality and efficiency. Comparative analysis against other scheduling algorithms showcases competitive performance, positioning MCTS as a promising solution for satellite task scheduling in dynamic environments.
\end{abstract}

\keywords{monte carlo tree search \and multi-satellite collection scheduling problem \and scheduling \and optimization}

\section{Introduction}
Every day, thousands of satellites orbit Earth conducting surveillance, capturing imagery, and relaying data around the globe. Due to the high cost of launch and the short lifespan of satellites, it is important to utilize this investment to its fullest potential. In order to do this, we must be able to schedule as many profitable tasks as possible during its lifetime. With a small number of well-spaced tasks, this is trivial. However as the number and frequency of tasks increase and weather and clouds create uncertainty as to the ability to complete tasks successfully, the problem becomes much harder. The multi-satellite collection scheduling problem (m-SatCSP) aims to optimize scheduling tasks over a collection of satellites under the uncertainty of cloud cover.

A wide range of approaches have been attempted for m-SatCSP, mainly focusing on finding an accurate heuristic. Many simplify the problem, removing things such as energy or memory constraints or the uncertainty of cloud cover. However, ignoring these constraints makes it hard for the solutions found in these papers to be applied to real-life situations where these constraints are present.

Monte Carlo Tree Search (MCTS) \cite{MCTS} is a search algorithm which uses random simulations to approximate the real value of making decisions. MCTS is frequently used in game-playing algorithms for board games such as Chess, Checkers, Othello, and Go as seen in AlphaGo \cite{AlphaGo}. Because MCTS plays out its simulations to the end of the game or scenario it is able to know the exact result of the given decisions, this means that with enough simulations it can approximate very closely the real-world values of each decision without using a heuristic to evaluate each position.

This paper explores using two versions of the Monte Carlo Tree Search algorithm to schedule satellites in m-SatCSP. We experiment with hyperparameters of the algorithm including the exploration coefficient and number of simulations to attempt to find the optimal configuration for speed and effectiveness. Once these values have been identified, the results of our algorithm will be compared against other algorithms in the literature to compare the quality and efficiency of its solutions. This paper is split into 3 parts: the background including previous work on this problem, a mathematical model, and graph formulation; a breakdown of the algorithms used including an explanation of MCTS and the modifications we made; and the experimentation and results section which includes our experimental setup and our results and analysis.


\section{Background}
Many different approaches have been taken to solve m-SatCSP varying widely in their success and speed. 

A dynamic programming approach to m-SatCSP was explored by Wang et al. in 2015 \cite{Wang}. In their process they broke the problem down by orbit, solving all possible combinations per orbit and combining them into a final path. They created and tested multiple different heuristics to evaluate the generated paths in order to select the best one. While this approach does show good results, as the number of orbits and tasks grows, the amount of processing time to achieve the desired result grows rapidly.

In their 2019 paper, Lam et al. \cite{TPNC} explore the use of advancements in Deep Q Learning \cite{DQN} to solve m-SatCSP. Their approach formulates each problem into a graph using structure2vec from Dai et al. \cite{struct2vec}, which is fed into a trained neural network. This network can take the graph and output the final path. The neural network is trained using reinforcement learning and so can learn from its actions over time to improve. Based on the results of the paper, it shows promise as due to its pretraining it is able to schedule in almost real-time. But, overall the results of tests fall short of scores achieved by \cite{Wang}.

Barkaoui and Berger \cite{Genetic} used a hybrid genetic algorithm to create a solution. They achieved this by transforming the m-SatCSP problem into a Vehicle Routing Problem with Time Windows (VRPTW) and using a hybrid genetic algorithm designed to solve VRPTW problems. The results reported by this paper are excellent, especially for large problems.

Chun et al. \cite{mdpiAEOSSP} uses a graph attention network-based decision neural network (GDNN) to solve the agile earth observation satellite scheduling problem (AEOSSP). They feed the current state of the scheduling graph into their trained GDNN and it selects each move taking into account a number of movement and time window constraints. The GDNN performed very well in the tests they conducted creating solutions matching some or exceeding the other algorithms and creating solutions extremely quickly due to its pre-training.

Recently, Peng et al. \cite{IEEETransitions} used a greedy randomized iterated local search (GRILS) algorithm to schedule agile earth observation satellites with time dependant transition times. Their procedure combines the greed randomised adaptive search procedure with an iterative local search to effectively schedule tasks with variable time windows. They two operators, INSERT and SHAKE, to create their schedules. The INSERT operator adds two unscheduled to the current schedule based on a greedy and randomness factor. The SHAKE operator removes a number of scheduled tasks in order to avoid getting stuck in local optimums due to greedy select. Their results are positive showing an average of between 70 to 85 percent of their expected values and short computing times.

Both Chen et al. \cite{MILP} and Song et al.\cite{MILP_Clouds} explore the use of a mixed-integer linear programming solution. By converting m-SatCSP to a mixed-integer linear programming problem, an LP solver can be employed to find a solution to that problem. \cite{MILP} focuses more on scheduling within a given time window rather than across the full window allowing more flexibility in the schedule. \cite{MILP_Clouds} looks at using MILP to schedule satellites under stochastic cloud cover similar to that of \cite{Wang}.

A Monte Carlo Tree Search algorithm was applied to satellite scheduling by Herrmann and Schaub \cite{MCTS_Sat} in order to schedule satellite tasks such as downlinking to ground stations, recharging, and collecting imagery. This differs from m-SatCSP which deals only with finding the optimal schedule for imagery tasks. In their problem, they used MCTS to approximate the Q-value of each action from their current state. In contrast, this paper will use MCTS to explore the decision tree to create a full task schedule over multiple orbits.


\subsection{Problem Formulation}
In order to easily compare with others, the mathematical model used by Wang et al. \cite{Wang} and Lam et al. \cite{TPNC} was used. The problem is broken down into a set of orbits $O$ and a set of tasks to be scheduled $T$. In order to avoid inter-orbit constraints, similar to \cite{Wang} and \cite{TPNC}, dummy tasks are added in between orbits. These tasks $s$ and $t$ denote the beginning and end of each orbit respectively. In m-SatCSP, the objective formula is as follows:

\begin{equation}
    max\sum_{i\in T} \varpi_i \biggl(1-\prod_{k\in O}\biggl(1-p_i^k\sum_{\substack{j \in T \\ j \neq i}} x_{i,j}^k\biggl)\biggl)
\label{ScoreCalc}
\end{equation}

where $\varpi_i$ is the amount of reward received from the completion of task $i$, $p_i^k$ is the probability that task $i$ will be successfully completed in orbit $k$, and $x_{i,j}^k$ is a binary value showing whether task $j$ is scheduled directly next to task $i$ in orbit $k$. Because we are not able to know if a run will be successful we take the probability of not completing the task due to cloud cover, $1-p_i^k$ for all scheduled instances and multiply them to get the chance that task $i$ is never completed. Taking one minus this probability gives us the probability of the task being completed which is then multiplied with the value of the task. The following constraints from \cite{Wang} are required:

\begin{equation}
   \forall k\in O, i \in T, \sum_{\substack{j\in T \cup \{t\} \\ j \neq i}} x_{ij}^k \leq b_i^k
   \label{schedulable}
\end{equation}

\begin{equation}
    \forall k\in O, i \in T, \sum_{\substack{j\in T\cup \{t\} \\ j \neq i}} x_{ij}^k = \sum_{\substack{j\in T\cup \{s\} \\ j \neq i}} x_{ji}^k
    \label{scheduled}
\end{equation}

\begin{equation}
    \forall k\in O, i \in T, x_{ij}^k = 0 \text{ if } we_i^k + st_{ij}^k > ws_j^k
    \label{windows}
\end{equation}

\begin{equation}
    \forall k \in O, \sum_{i \in T}\sum_{\substack{j \in T \cup \{t\} \\ j \neq i}} x_{ij}^k (we_i^k - ws_i^k)m_k \leq M_k
    \label{memory}
\end{equation}

\begin{equation}
    \forall k \in O, \sum_{i \in T\cup \{t\}}\sum_{\substack{j \in T \\ j \neq i}} x_{ij}^k (we_i^k - ws_i^k)e_k + \sum_{i \in T}\sum_{\substack{j \in T \\ j \neq i}} x_{ij}^k \rho_{ij}^k \leq E_k
    \label{energy}
\end{equation}

The notation for these equations can be found in Table \ref{table:notation}. Equation \eqref{schedulable} ensures that task $i$ is only scheduled in orbits when it is available. Equation \eqref{scheduled} ensures that if a task is scheduled it has a defined predecessor and successor and that if it is not scheduled then it does not. Equation \eqref{windows} ensures that if a task is scheduled then the end time of the previous task and the amount of time needed to slew between the two tasks finishes before the start of the next task. Equation \eqref{memory} ensures that the amount of memory used by scheduled tasks in orbit $k$ does not exceed the maximum memory usage allowed for that orbit. Equation \eqref{energy} ensures that the amount of energy used to complete each scheduled task and the amount of slew energy required between the scheduled tasks does not exceed the maximum energy use allowed for orbit $k$.

\begin{table}[htbp]
\caption{Notation (from Lam et al. \cite{TPNC})}
\label{table:notation}
\centering
\begin{tabular}{lp{2.5in}}
\toprule
Symbol & Description\\
\midrule
$O$  & set of orbits, $O = \{1,..,M\}$       \\
$T$  & set of tasks, $T = \{1,..,n\}$        \\
$i,j$  & task index, $i,j \in T$             \\
$k$  & orbit index, $k \in O$                \\
$\varpi$  & profit of task $i$, $i \in T$    \\
$p_i^k$  & probability that task $i$ will be observed during orbit $k$, $i \in T, k \in O$ \\
$x_{ij}^k$  & $x_{ij}^k$ is 1 if task $j$ is scheduled after task $i$ in orbit $k$, $i,j \in T, k \in O$ \\
$b_{i}^k$  & $b_{i}^k$ is 1 if task $i$ can be scheduled in orbit $k$, $i \in T, k \in O$ \\
$M_k, E_k$  & memory and energy capacity for orbit $k$, $k \in O$ \\
$m_k, e_k$  & memory and energy consumption per unit of time in orbit $k$, $k \in O$  \\
$[ws_i^k, we_i^k]$ & observation time window for task $i$ in orbit $k$, $i \in T, k \in O$ \\
$st_{ij}^k$ & setup time between task $i$ and task $j$ in orbit $k$, $i,j \in T, k \in O$ \\
$\rho_{ij}^k$ & slew energy required between task $i$ and task $j$ in orbit $k$, $i,j \in T, k \in O$ \\ \hline
\end{tabular}
\end{table}


\subsection{Graph Formulation}
\begin{figure}[htbp]
\centering
    \includegraphics[width=\textwidth]{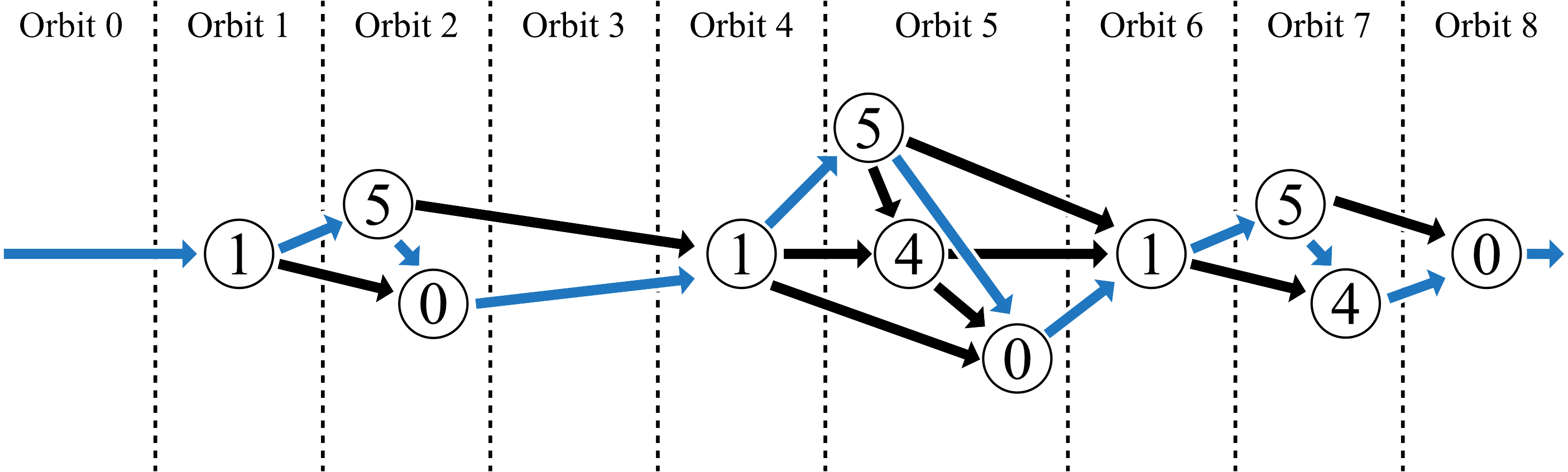}
    \caption{An example problem graph for Wang \cite{Wang} 9 Orbit, 10 Task, Problem 4. In this example, all nodes are connected, but this is not always the case. Not all paths through the graph are possible, for example, in orbit 5 the path 5-4-0 is not possible due to energy constraints. The best path for this problem is highlighted in blue. }
    
    \label{fig:example-graph}
\end{figure}

In order to trace paths using MCTS, the m-SatCSP problems were converted into a directed graph. This graph was split by orbit with each orbit beginning with a mode $s^k$ and ending with a node $t^k$. The node $t^k$ is equivalent to $s^{k+1}$ when reconstructing the full multi-orbit graph except for $s^1$ and $t^M$ which are placed at the beginning and end of the problem respectively. Each vertex in the graph represents a task available in the orbit $k$ which is regulated by Equation \eqref{schedulable}. The connection between vertices is controlled by Equation \eqref{windows}, ensuring that there is enough transition time between the two nodes to allow for slewing. Since the $s$ and $t$ nodes do not serve any operational requirements (they only mark orbit transition), there are no transition time, slewing requirements, or memory and energy requirements associated with these nodes. Graph traversal constraints regarding energy and memory usage are calculated during the scheduling process as they depend on the previously selected nodes when traversing the graph. An example of a problem graph can be seen in Figure \ref{fig:example-graph}.

\section{Algorithms}
\subsection{Monte Carlo Tree Search}
Monte Carlo Tree Search is an algorithm which builds a tree based on random simulations through a set of states. Each simulation allows the tree to expand, but rather than expanding evenly as a regular tree search algorithm such as minimax does, MCTS expands the tree unequally favouring paths which have shown promising results. With a large enough number of simulations, MCTS is able to approximate the real score acquired by making each decision along a path. In this paper, two versions of the Monte Carlo Tree Search algorithm were used, the first is a standard version which holds the average value of all paths reached from that node and the second only retains the best value achieved from that node.

\begin{figure}[htbp]
    \centering
    \includegraphics[width=\columnwidth]{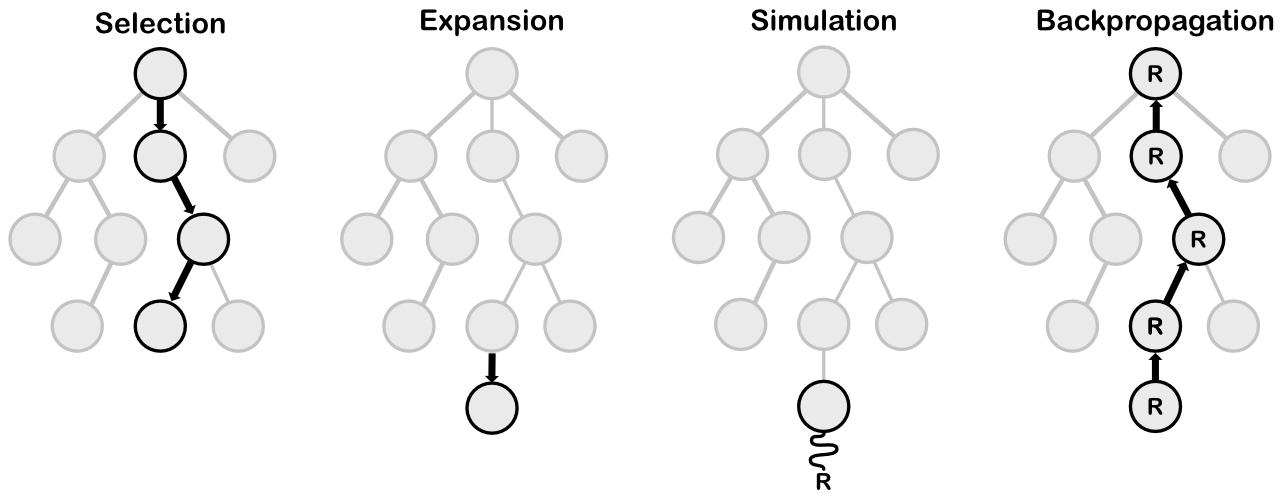}
    \caption{\centering A visual representation of the phases of Monte Carlo Tree Search}
    \label{fig:mcts-phases}
\end{figure}

The MCTS algorithm takes place in four phases: Selection, Expansion, Simulation, and Backpropagation. A visual example of these phases can be seen in Figure \ref{fig:mcts-phases}.

\subsubsection{Selection}
The first phase of the Monte Carlo Tree Search algorithm is selection. In this phase, the tree is traversed, and a node is selected to expand. This selection process begins at the root of the tree, which in this case is an empty node representing the beginning of orbit zero. From there it proceeds down the tree by selecting nodes until it reaches a leaf node. In order to select a node, the UCT (Upper Confidence Bound applied to Trees) formula is used. UCT selects the next node based on its value and the amount it has been previously explored and is defined as the following:

\begin{equation}
\frac{w_{i}}{n_{i}}+c\sqrt{\frac{\ln{N_i}}{n_i}}
\label{UCT}
\end{equation}

where $w_i$ is the cumulative value of the node $i$, $n_i$ is the number of times that node has been visited, $N_i$ is the number of times the parent of $i$ has been visited, and $c$ is the exploration coefficient. The exploration coefficient allows for the modification of the amount of exploration done by the algorithm. The higher the exploration is the more likely it is to choose to expand all paths more evenly leading to a fuller tree. Conversely, if the exploration coefficient is small then the algorithm is more likely to focus only on the best branches of the tree found. Balancing this coefficient is important to get the best results out of the MCTS algorithm, if set too small it may miss the best solution which could be hidden down a path which may seem non-optimal at first but set too large and it may waste simulations growing the complete tree similarly to other tree searches. 

The UCT calculation is computed for all children of the current node and the node with the highest value is selected as the candidate. This process is repeated until a leaf node is reached.

\subsubsection{Expansion}
Once a leaf node is reached, the children of the leaf node are added to the tree. Because none of the children have been visited yet, the UCT value can not be calculated for them and so one of them is selected randomly to be simulated. On recurring selections, if a node hasn't been visited it will be assigned a UCT value of positive infinity, simply bumping it to the top of the list of nodes to be selected. If a node reached by selection has no children to be expanded (i.e. a terminal node) then the simulation stage is skipped, and the value of the node is backpropagated. The only nodes considered to be terminal in this paper are all tasks in the final orbit as they can be the last task completed in a given schedule.
\subsubsection{Simulation}
Once a node is selected to be simulated, a simulation is run beginning in the state of the selected node. From that state, decisions are made at random, giving no bearing to their value, until a terminal state is reached. Once a terminal state has been reached, the final value of the path is calculated and passed on to be backpropagated. Simulation types other than random simulation can be used, including low-depth minimax algorithms in order to select paths in the simulation which better outcomes but this can slow down the simulation process such that you will obtain fewer simulations in the same amount of time.
\subsubsection{Backpropagation}
In the final phase of the MCTS algorithm, the value found in the simulation is added to the value of each of the nodes on the path through the tree and their visit counts are each increased by one. This update allows the algorithm to increase its understanding of the search space over time and to understand the expected value of each decision.

\subsubsection{Final Selection}
Once all simulations are completed, a final selection must take place to decide what moves to make. In a normal application of MCTS such as in a board game, this selection would only happen from the root to select the next move made by the current player. This is done by selecting the move from the root with the greatest number of visits and thus the move that the algorithm feels has the most promise. In this paper, the algorithm is used to generate the complete path of scheduled tasks and so this selection process is used to select nodes until a terminal node is reached. This path is then able to have its score calculated using Equation \eqref{ScoreCalc}. 

\subsection{Alterations}
In this paper, two different versions of MCTS were used. The first is the standard version outlined above. The second was modified to keep the best value found from that node rather than the average score. This change slightly modifies some of the steps of the algorithm, namely the selection and backpropagation steps. In the selection step the UCT calculation is modified slightly to account for this change in the value of each node:

\begin{equation}
W_{i}+c\sqrt{\frac{\ln{N_i}}{n_i}}
\label{UCT_mod}
\end{equation}

with $W_i$ now being the maximum value found from a given node during all simulations. The other change comes in backpropagation where instead of simply adding the value found by the rollout simulation, the value is compared to the current value of each node and replaced only if it is greater than the previously found value. Finally, when selecting the final path to take, it becomes much easier as we just follow the path of nodes with the highest value.


\section{Experiments and Results}

\subsection{Method}
The aim of this paper is to evaluate and demonstrate the capability of Monte Carlo Tree Search to efficiently and effectively schedule satellite tasks under mSat-CSP. In order for MCTS to perform effectively we must first fine-tune its hyperparameters for it to perform effectively. Once the effective hyperparameters have been found, we will compare the MCTS algorithms to other models found in current literature, comparing them in terms of time taken to form a solution and the value of the solution based on Equation \eqref{ScoreCalc}.

In this paper, two experiments were conducted to test which of the two versions of the MCTS algorithm demonstrated better performance and to evaluate what hyperparameters achieved the best performance. The two hyperparameters that were tested were the value of the exploration coefficient seen in Equation \eqref{UCT} and Equation \eqref{UCT_mod} and the number of simulations that each algorithm completed before the path found was evaluated. In each of these experiments, the score achieved by the algorithm, calculated using Equation \eqref{ScoreCalc}, and the time taken to complete the problem were used as metrics to judge their success. Each problem was run three times and the value and time taken were averaged over these three trials.  After these tests were evaluated, the best-performing combination was tested on more test sets and compared to results found in \cite{Wang}, \cite{TPNC}, and \cite{Genetic}. The expected values shown in the tables are based on the values found by Heuristic 5 in \cite{Wang} for the 9, 21, and 42 orbit test sets and from the Hybrid Genetic Algorithm in \cite{Genetic} for the 300 and 500 tasks 15 orbit test sets.

\subsubsection{Test Sets}
The test sets used for these experiments were created and used by Wang in \cite{Wang}. There are 15 different test sets created by Wang which range in number of orbits and number of tasks. The test sets are referred to by their number of orbits and number of tasks in the following fashion, \mbox{\# Orbits-\# Tasks}. For example, 9-10 refers to the test set with 9 orbits and 10 tasks. Each test set contains 10 different problems in that combination of orbits and tasks. The number of orbits included are 9, 21, and 42 and the number of tasks included are 10, 20, 30, 40, 50, 60, 120, 180, and 200. The experiments conducted only use the small 9 orbit test sets (10, 20, 30, 40, 50, 60 Tasks), but the final comparison contains all data sets from \cite{Wang}.

In addition to the test sets from \cite{Wang}, the larger test sets created by \cite{Genetic} were also used. There are two sets of tasks from \cite{Genetic}, both with 15 orbits, one with 300 tasks and the other with 500 tasks. These test sets are built in the same format as the ones above but are much more dense, containing more potential paths and more tasks overall.  

\subsubsection{Hardware/Software}
All experiments were run on a Ryzen 7 7800x @ 3.8GHz CPU with 32GB of memory. All code was written in Java and run using the IntelliJ IDEA IDE.


\subsection{Results}

\subsubsection{Evaluation Based on Exploration Coefficient}
Table \ref{table:explore} provides an overview of the performance of the Average and Max MCTS algorithms using differing exploration coefficients. The exploration coefficient values tested were 1.5, 5, and 10, with a higher value increasing the odds for the algorithm to select a less-explored path, thus expanding the tree. Each of the test sets in Table \ref{table:explore} shows the average value achieved over all 10 problems in the set with three tests run on each problem. 

Based on the results in this table, it can be seen that, in general, the max algorithm shows superior performance compared to the average algorithm. All of these experiments were run for 10 million simulations before the best path was scored. 

\begin{table}[htbp]
\centering
\caption{Performance Comparison of Exploration Coefficients on 9-Orbit Data Sets: Average Achieved Values Across Three Runs}
\begin{tabularx}{\textwidth}{ccCCCCCC}
\toprule
 \multirow{2}{*}{\begin{tabular}[C]{@{}l@{}} \textbf{Problem} \textbf{Set} \end{tabular}} &
  \multirow{2}{*}{\begin{tabular}[C]{@{}l@{}}\textbf{Expected}  \textbf{Value}\end{tabular}} &
  \multicolumn{3}{c}{\textbf{Max Algorithm}} &
  \multicolumn{3}{c}{\textbf{Average Algorithm}} \\ 
   &
   &
  \textbf{1.5} &
  \textbf{5} &
  \textbf{10} &
  \textbf{1.5} &
  \textbf{5} &
  \textbf{10} \\ \midrule
    9-10 & 12.55 & 12.54 & 12.30 & 12.54 & 12.54 & 12.51 & 12.51 \\
    9-20 & 26.33 & 25.71 & 26.33 & 26.33 & 26.00 & 25.91 & 26.14 \\
    9-30 & 28.86 & 28.27 & 28.82 & 28.82 & 28.17 & 28.32 & 28.10 \\
    9-40 & 50.60 & 45.83 & 48.03 & 49.36 & 46.56 & 48.63 & 45.84 \\
    9-50 & 55.24 & 48.37 & 52.74 & 54.03 & 51.08 & 52.81 & 53.63 \\
    9-60 & 59.01 & 52.63 & 57.03 & 58.98 & 54.91 & 56.00 & 56.43 \\ \midrule
    Average & 38.77 & 35.56 & 37.54 & 38.34 & 36.54 & 37.36 & 37.11 \\ \midrule
\multicolumn{2}{c}{Average Time (s)} & 9.54  & 9.85  & 9.99  & 9.74  & 9.43  & 9.72  \\
\bottomrule
\end{tabularx}
\label{table:explore}
\end{table}

\subsubsection{Evaluation Based on the Number of Simulations Run}
For this experiment, each algorithm was tested using three different numbers of simulations: 1 million, 5 million, and 10 million simulations. The results for each of these tests on the 9 orbit test set can be found in Table \ref{table:sims}. 

Generally, the higher the number of simulations the better the final value achieved, with the max algorithm achieving on average 99.0\% of the expected value as compared to the 97.6\% for the average algorithm over the 9 orbit test set. This outcome is to be expected because, as the number of simulations is increased the tree expands more and thus it is more likely that the path to the best solution will be discovered.

\begin{figure}[htbp]
    \centering
    \includegraphics[width=\columnwidth]{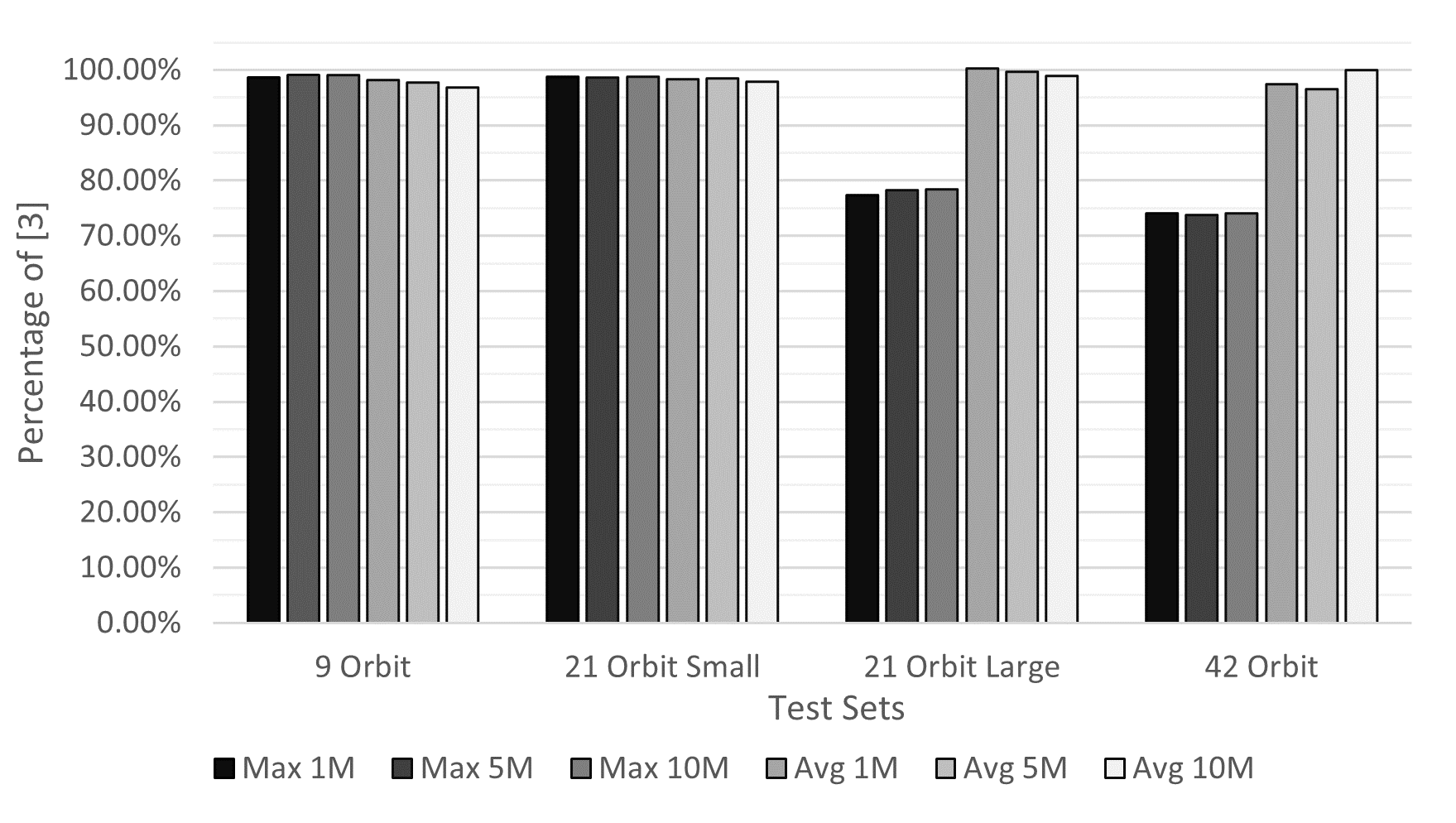}
    \caption{A graphical representation of the performance of each algorithm on all test sets from \cite{Wang}.}
    \label{fig:percentage-graph-wang}
\end{figure}

\begin{figure}[htbp]
    \centering
    \includegraphics[width=\columnwidth]{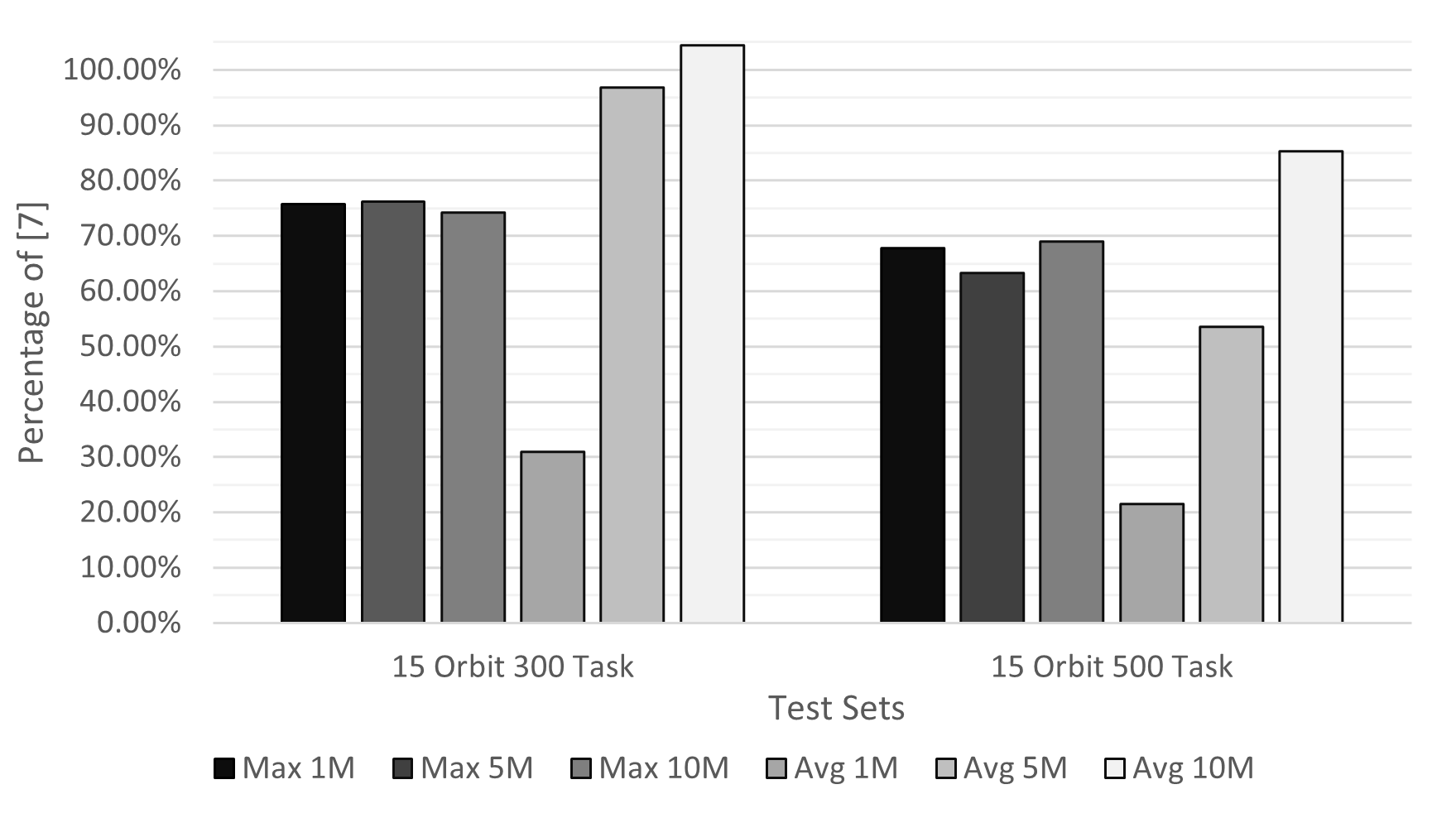}
    \caption{A graphical representation of the performance of each algorithm on all test sets from \cite{Genetic}.}
    \label{fig:percentage-graph-bark}
\end{figure}

When looking at the results over all of the test sets from \cite{Wang}, shown in Table \ref{table:sims-all}, a different outcome can be seen. While the max algorithm performs slightly better over the test sets with small numbers of tasks, the average algorithm performs vastly better over the increased orbit and increased task load sets, even finding better solutions than Wang \cite{Wang} on some problems. The average algorithm is also faster than the max algorithm. A visual representation of the MCTS algorithms' performance against Wang's test sets can be seen in Figure \ref{fig:percentage-graph-wang} with the values from \cite{Wang} being 100 percent. Similarly, the performance of MCTS against the test sets from \cite{Genetic} can be seen in Figure \ref{fig:percentage-graph-bark} using values from \cite{Genetic} as the 100 percent baseline.

\begin{table}[htbp]
\centering
\caption{Performance Comparison of Number of Simulations on 9-Orbit Data Sets: Average Achieved Values Across Three Runs}
\begin{tabularx}{\textwidth}{ccCCCCCC}
\toprule
 \multirow{2}{*}{\begin{tabular}[c]{@{}l@{}} \textbf{Problem} \textbf{Set} \end{tabular}} &
  \multirow{2}{*}{\begin{tabular}[c]{@{}l@{}}\textbf{Expected} \textbf{Value}\end{tabular}} &
  \multicolumn{3}{c}{\textbf{Max Algorithm}} &
  \multicolumn{3}{c}{\textbf{Average Algorithm}} \\ 
   &
   &
  \multicolumn{1}{c}{\textbf{1.5M}} &
  \multicolumn{1}{c}{\textbf{5M}} &
  \textbf{10M} &
  \multicolumn{1}{c}{\textbf{1.5M}} &
  \multicolumn{1}{c}{\textbf{5M}} &
  \textbf{10M} \\ \midrule
    9-10 & 12.55 & 12.54 & 12.54 & 12.54 & 12.54 & 12.51 & 15.51 \\
    9-20 & 26.33 & 26.33 & 26.33 & 26.33 & 26.25 & 26.15 & 26.14 \\
    9-30 & 28.86 & 28.85 & 28.85 & 28.82 & 28.55 & 28.24 & 28.10 \\
    9-40 & 50.60 & 48.87 & 49.49 & 49.36 & 49.04 & 49.24 & 45.84 \\
    9-50 & 55.24 & 53.90 & 54.44 & 54.03 & 53.80 & 53.88 & 53.63 \\
    9-60 & 59.01 & 57.86 & 58.09 & 58.98 & 57.26 & 56.70 & 56.43 \\ \midrule
    Average & 38.77 & 38.06 & 38.29 & 38.34 & 37.91 & 37.79 & 37.11 \\ \midrule
    \multicolumn{2}{c}{Average Time (s)} & 1.00  & 5.10  & 9.99  & 0.99  & 4.98  & 9.72  \\ \bottomrule
\end{tabularx}
\label{table:sims}
\end{table}

\begin{table}[htbp]
\caption{Reward and Time for All Test Sets Used in Wang \cite{Wang} Averaged Over Three Runs}
\centering
\begin{tabularx}{\textwidth}{cccCCCCCC} 
\toprule
\multicolumn{2}{c}{\multirow{2}{*}{\begin{tabular}[c]{@{}l@{}}\textbf{Problem Set}\end{tabular}}} & \multirow{2}{*}{\begin{tabular}[c]{@{}l@{}} \textbf{Expected Value} \end{tabular}} & \multicolumn{3}{c}{\textbf{Max Algorithm}}                                    & \multicolumn{3}{c}{\textbf{Average Algorithm}}                                 \\ 
\multicolumn{2}{c}{}                                                                        &                                                                            & \textbf{1M}      & \textbf{5M}     & \textbf{10M}     & \textbf{1M}     & \textbf{5M}      & \textbf{10M}      \\ 
\midrule
\multirow{2}{*}{9 Orbits}                                                      & Value        & 38.77                                                                      & 38.06                        & 38.29                        & 38.34   & 37.91                        & 37.79                        & 37.11    \\
                                                                               & Time         & N/A                                                                        & 1.00                         & 5.10                         & 9.99    & 0.99                         & 4.98                         & 9.72     \\ 
\midrule
\multirow{2}{*}{\begin{tabular}[c]{@{}l@{}}21 Orbits\\ Small\end{tabular}}     & Value        & 61.04                                                                      & 59.93                        & 60.09                        & 59.93   & 59.78                        & 59.88                        & 59.55    \\
                                                                               & Time         & N/A                                                                        & 7.36                         & 7.58                         & 15.18   & 1.48                         & 7.37                         & 14.64    \\ 
\midrule
\multirow{2}{*}{\begin{tabular}[c]{@{}l@{}}21 Orbits\\ Large\end{tabular}}     & Value        & 285.75                                                                     & 221.00                       & 224.01                       & 224.22  & 287.39                       & 285.62                       & 285.12   \\
                                                                               & Time         & N/A                                                                        & 10.61                        & 51.13                        & 122.00  & 6.93                         & 30.96                        & 97.29    \\ 
\midrule
\multirow{2}{*}{\begin{tabular}[c]{@{}l@{}}42 Orbits\\ Large\end{tabular}}     & Value        & 498.37                                                                     & 366.88                       & 364.18                       & 365.90  & 484.06                       & 480.10                       & 480.55   \\
                                                                               & Time         & N/A                                                                        & 18.65 \`{}                   & 91.77                        & 373.05  & 12.87                        & 60.60                        & 121.90   \\ 
\midrule
\multirow{2}{*}{\begin{tabular}[c]{@{}l@{}}15 Orbits\\ 300 Tasks\end{tabular}} & Value        & 630.20                                                                     & 477.27                       & 479.91                       & 468.02  & 194.68                       & 609.97                       & 658.39   \\
                                                                               & Time         & N/A                                                                        & 180.24                       & 895.49                       & 1740.32 & 222.45                       & 916.40                       & 1747.74  \\ 
\midrule
\multirow{2}{*}{\begin{tabular}[c]{@{}l@{}}15 Orbits\\ 500 Tasks\end{tabular}} & Value        & 756.26                                                                     & 512.18                       & 521.11                       & 521.99  & 163.05                       & 404.86                       & 645.26   \\
                                                                               & Time         & N/A                                                                        & 313.85                       & 1547.11                      & 3075.69 & 577.38                       & 1532.48                      & 2673.24  \\ 
\midrule
\multicolumn{3}{c}{Average Percentage of Expected Value}                                                                                                                                   & 82.08\% & 81.54\% & 82.28\% & 74.47\% & 90.49\% & 97.24\%  \\
\bottomrule
\end{tabularx}
\label{table:sims-all}
\end{table}

\subsection{Discussion}

In order to accurately access the performance of the algorithm outlined in this paper, it will be compared not only to the values found by \cite{Wang} but also to other algorithms using the same set of test sets. The other types of algorithms that will be compared against are Deep-Q Learning using scores from \cite{TPNC} as well as a Modified Genetic Algorithm from \cite{Genetic}.

These Algorithms will be compared across the whole spectrum of test sets available from \cite{Wang} and \cite{Genetic}. The best achieved average values for each test set will be chosen for each algorithm, the values of which can be found in Table \ref{table:compare}. 

\begin{table}[htbp]
\centering
\caption{Reward for Wang \cite{Wang} and Barkaoui \cite{Genetic} Test Sets for Different Algorithms}
\begin{tabularx}{\textwidth}{cCCCC}
\toprule
\textbf{Problem Set}        & \textbf{DP \cite{Wang}} & \textbf{MCTS (Best)}   & \textbf{DQN \cite{TPNC}}    & \textbf{GA \cite{Genetic}}\\ \midrule
9 Orbits           & \textbf{38.77}  & 38.34   & 32.73   &  N/A  \\ 
21 Orbits Small    & \textbf{61.04}  & 60.09   & 54.37   &  N/A  \\ 
21 Orbits Large    & 285.75 & \textbf{287.39}  & 186.88  &  N/A  \\ 
42 Orbits Large    & 498.37 & 484.06  & N/A     &  \textbf{673.19}  \\ 
15 Orbits 300 Tasks & N/A   & \textbf{658.39}  & N/A   & 630.20  \\ 
15 Orbits 500 Tasks & N/A   &  645.26 & N/A     &   \textbf{756.26} \\ \bottomrule
\end{tabularx}
\label{table:compare}
\end{table}

\begin{table}[htbp]
\centering
\caption{Time for Wang \cite{Wang} and Barkaoui \cite{Genetic} Test Sets for Different Algorithms in seconds}
\begin{tabularx}{\textwidth}{cCCCC}
\toprule
\textbf{Problem Set}        & \textbf{DP \cite{Wang}} & \textbf{MCTS (Best)}   & \textbf{DQN \cite{TPNC}}    & \textbf{GA \cite{Genetic}}\\ \midrule
9 Orbits           & \textbf{0.0015}  & 9.99   & 0.30  &  N/A  \\ 
21 Orbits Small    & \textbf{0.004}  & 7.58   & 0.60   &  N/A  \\ 
21 Orbits Large    & 60.00 & \textbf{6.53}  & 7.20  &  N/A  \\ 
42 Orbits Large    & 60.00 & \textbf{12.87}  & N/A     &  1800.00  \\ 
15 Orbits 300 Tasks & N/A  & \textbf{1747.74}& N/A   & 1800.00   \\ 
15 Orbits 500 Tasks & N/A   &  2673.24       & N/A     & \textbf{1800.00}  \\ \bottomrule
\end{tabularx}
\label{table:compare-time}
\end{table}

On the smaller orbit sets from \cite{Wang}, the MCTS algorithm performs almost as well as the dynamic programming approach from \cite{Wang} reaching within one percent of Wang's values. The MCTS algorithm also performs far better than the Deep Q-Learning algorithm from \cite{TPNC} in terms of the quality of the result. However, when looking at the time taken to get these results in Table \ref{table:compare-time}, due to its fixed amount of simulations, MCTS takes much more time as it must complete all of the required simulations even if the optimal solution is found early. When comparing the MCTS algorithms, better results were achieved on these problems with the max algorithm.

When moving on to the larger test sets created by \cite{Wang} because of time constraints implemented by Wang, the dynamic programming heuristic was not always able to find the optimal solution for the larger problems. These test sets are where the MCTS algorithm really worked well, specifically the average algorithm which was able to find solutions extremely efficiently when compared to its DP and DQN counterparts. While the average MCTS algorithm did quite well, the max algorithm struggled a little. We believe this is due to the modification made to the UCT calculation in Equation \eqref{UCT_mod}. In this category MCTS one not only on speed but also on quality with average MCTS algorithm achieving an almost five percent increase of the values found by \cite{Wang} in the 21 orbit 200 tasks set. In terms of time, MCTS was better than both the DP and DQN solutions for both large test sets. While the Genetic algorithm from \cite{Genetic} did get better results, it was also run for 1500 times the amount of time, so this is to be expected.

The test sets from Barkaoui \cite{Genetic} were considerably denser and thus more complex to schedule both due to their increased number of tasks nd also the number of time windows and how they overlap. For the 300 task set, MCTS was always faster than the GA solution as Barkaoui and Berger always let their algorithm run for 30 minutes no matter what. With 1 million simulations the average algorithm found it challenging to be able to explore all the complexity of these larger data sets leading to a poor performance. However, the 5 million and 10 million simulation runs performed quite well with the 10 million simulation run achieving a 5 percent increase overall and over a 20 percent increase on one of the problems averaged over three runs (one run of the problem getting upwards of a 29 percent increase).

When put up against the 500 task test set from \cite{Genetic}, due to the extreme density of the problem set, our MCTS algorithms did not fair as well as with the other tests. For this test set the best outcome came from the average algorithm with 10 million simulations though this only achieved 86 percent of the value achieved by \cite{Genetic} and took 1.5 times as long to complete each problem.


\section{Conclusion}

In conclusion, this paper has explored the multi-satellite collection scheduling problem (m-SatCSP) and proposed the use of Monte Carlo Tree Search (MCTS) as an effective algorithmic framework for addressing this complex optimization problem. After performing hyperparameter optimization it was found that generally a higher number of simulations and a higher exploration coefficient yielded better results.

Overall, the MCTS algorithm generally performed at the same level or better than the other algorithms discussed in both the speed and quality of the results. While the solutions created by Lam et al. using Deep Q-Learning \cite{TPNC} were extremely time efficient, they were not as good as the other solutions in the amount of profit attained. Wang's dynamic programming solution \cite{Wang} was able to generate comparable results to our MCTS solution but took much more time on the larger more complex test sets. The hybrid genetic algorithm solution from Barkaoui and Berger \cite{Genetic} was able to generate impressive solutions to larger problems but was generally quite slow due to its fixed run time.

Both the strengths and weaknesses of the MCTS algorithm discussed here lie in its stochasticity. It allows the algorithm to potentially find solutions very quickly but also makes it inconsistent in finding solutions in more dense and challenging situations. With more optimization and testing, Monte Carlo Tree Search could be an excellent contender for use as a scheduling algorithm.


\bibliographystyle{unsrt}  
\bibliography{references}

\end{document}